\documentclass{article} 
\usepackage{iclr2020_conference,times}

\usepackage{amsmath,amsfonts,bm}

\def\eqref#1{equation~\ref{#1}}

\def\1{\bm{1}}

\DeclareMathAlphabet{\mathsfit}{\encodingdefault}{\sfdefault}{m}{sl}
\SetMathAlphabet{\mathsfit}{bold}{\encodingdefault}{\sfdefault}{bx}{n}

\usepackage{hyperref}
\usepackage{url}

 \usepackage{booktabs}
 \usepackage{arydshln}
 \usepackage{colortbl}

\title{Unsupervised Sentiment Analysis for Code-Mixed Data}

\author{Siddharth Yadav, Tanmoy Chakraborty 
\\
IIIT-Delhi, India \\
\texttt{\{siddharth16268,tanmoy\}@iiitd.ac.in} \\
}

\iclrfinalcopy 
\begin{document}

\maketitle

\begin{abstract}

Linguistic code-mixing or code-switching is the practice of alternating between two or more languages. Mostly observed in multilingual societies, its occurrence is increasing and therefore its importance. A major part of sentiment analysis research has been monolingual and best of the monolingual methods perform poorly on code-mixed text. 
In this work, we introduce methods that use different kinds of multilingual and crosslingual embeddings to efficiently transfer knowledge from monolingual text to code-mixed text for sentiment analysis of code-mixed text. Our methods can handle code-mixed text in a zero-shot way quite well. Our methods beat state-of-the-art on English Spanish code-mixed sentiment analysis by an absolute 3\% F1 score. We are able to achieve a F1 score of 0.58 (without parallel corpus) and of 0.62 (with parallel corpus) on the same benchmark in a zero-shot way as compared to 0.68 F1 score in supervised settings. Our code is publicly available\footnote{\url{https://github.com/sedflix/unsacmt}} 

\end{abstract}

\section{Introduction}

Code-mixing or code-switching is the practice of alternating between two or more languages in a conversation. Mostly observed in multilingual societies like Europe and India, it is often associated with informal or casual conversation/speech like in social media or chats or in-person talks. Its occurrence is increasing rapidly and hence its importance. In fact, code-mixing is the default mode of communication in many places like some part of India and Mexico \citep{PARSHAD2016375}. Often, bilingual and multilingual people talk in a code-mixed language because it puts less cognitive load on their brain. People can use words or phrases from either languages, whichever comes first to their mind. In other words, it's more constrictive to think in a monolingual way. Most of the NLP system in today's world are monolingual. \cite{Choudhury2007} shows language used on these social media platforms, from where we get most of the code-mixed text, differs from the standard language that is found in more formal texts like books. Monolingual NLP methods perform poorly on code-mixed data. 

Therefore, it is crucial to build technology for code-mixed text. Building this technology comes with its own challenges as described in \cite{cetinoglu-etal-2016-challenges}. Summarising it, analysis of code-mixed text is hard due to lack of code-mixed text corpus and datasets, large amount of unseen constructions caused by combining lexicon and syntax of two or more language, and large number of possible combinations of code mixed languages 
Constructing one dataset requires lot of manual labour and construction of datasets for each code-mixed pair seems infeasible. Hence, it will be more beneficial to build technology that doesn't require code-mixed dataset and is able to transfer knowledge from related monolingual datasets to code-mixed domain efficiently. In other words, it would be beneficial if our technology is unsupervised.   

In this work, we introduce methods that use different kinds of multilingual and crosslingual embeddings to efficiently transfer knowledge from monolingual text to code-mixed text for sentiment analysis of code-mixed text. We make our methods independent of any kind supervised signals between two languages by not using any kind of Machine Translation or Language Identification systems, as these systems perform poorly on low-resource language pair due to lack of proper datasets. Our goal in this work to find out embedding spaces and training techniques that assist efficient knowledge transfer from monolingual text to code-mixed text. 

Crosslingual/multilingual embeddings are an obvious starting point for our goal. There is a subtle difference between crosslingual and multilingual embeddings. In crosslingual world, we are aware of the language of the given data point. Therefore, we can choose an appropriate model or a part of the model corresponding to that language. In multilingual, we not aware of the language of the given data point. Therefore, multilingual models are totally language agnostic. Regardless the difference, crosslingual and multilingual embeddings implies a shared vector space among all the involved languages. There promise is that once a model is trained for any one language, the model can do predictions on any other language. When using crosslingual embeddings, the languages of data points should be known beforehand. \cite{conneau2017word} has shown that crosslingual embeddings can be made in a totally unsupervised way, i.e. they only require monolingual embeddings of the respective languages and making monolingual embeddings only require monolingual corpus which is easily available for even low resource languages, and these embeddings perform at par to supervised crosslingual embeddings. This helps our goal of being independent of any kind supervised signals between two languages. 

As sentence level crosslingual embeddings can't handle two different languages in a single sentence or document, only word/subword/character level crosslingual embeddings can be used for code-mixed text. All kinds of multilingual embeddings are much better suited for analysing code-mixed text because they don't require language tags and we are not aware of the language tag of each word in code-mixed text. We can merge vocabulary(and embeddings) of word/subword level crosslingual embeddings to create a shared vocabulary resulting in a the new embedding that can handle occurrence of different languages in a single data point without the need of language tags. Just by using these crosslingual/multilingual embeddings in vanilla classifiers, we can train the classifier on available monolingual datasets and do predictions on code-mixed text. This gives a way to analysis code-mixed text in a zero-shot way. When used in supervised settings, the downstream LSTM layers can capture semantic structure specific to code-mixed text over those embeddings.

It is important to note that crosslingual and multilingual embeddings are trained on monolingual sentences. Code-mixed text has it own large set of syntactic structures and semantic association which are not captured in monolingual sentences. Therefore, crosslingual and multilingual embeddings are not ideal for learning code-mixed text analysis. Embedding for code-mixed text should utilised code-mixed text to capture the fine syntactic and semantic features of code-mixed text \citep{pratapa-etal-2018-word}.




\section{Related Work}

\cite{en_es_intro} was the first work to introduce sentiment analysis on English-Spanish code-mixed data. 
\cite{pratapa-etal-2018-word} focuses on embeddings for improving the downstream tasks on English-Spanish code-mixed data. They show that using pretrained embeddings learned from code-mixed data perform better than bilingual embeddings, like Bilingual correlation based embeddings(BiCCA) \citep{ faruqui-dyer-2014-improving}, Bilingual compositional model(BiCVM) \citep{hermann-blunsom-2014-multilingual} and Bilingual Skip-gram (BiSkip) \citep{luong-etal-2015-bilingual}. They propose a new approach of training skip-grams on synthetic code-mixed text generated approach mentioned by \citep{pratapa-etal-2018-language} which requires large parallel corpus. 
\cite{samanta-etal-2019-improved} augments monolingual datasets using Machine Translation systems and their own code-mixed generative models to assist code-mixed classification.
Genta Indra Winata proposed Multilingual Meta-Embeddings \citep{winata-etal-2019-hierarchical} and Hierarchical Meta-Embeddings \citep{winata-etal-2019-learning} which combine different kinds of pretrained monolingual embeddings at word, subword, and character-level into a single language-agnostic lexical representation without using any specific language identifiers.

There have been some interesting work in proposing new models for sentiment analysis English-Hindi code-mixed text. \cite{sharma_2015} proposes an approach based on lexicon lookup for text normalization. 
\cite{Pravalika_2017} used a lexicon lookup approach to perform domain specific sentiment analysis. 
\cite{joshi-etal-2016-towards} uses sub-words with CNN-LSTMs to capture sentiment at morpheme level. 
\cite{en_hi_pre_sota} uses Siamese networks to map code-mixed and standard language text to a common sentiment space.
\cite{en_hi_stoa} proposes use of Dual Encoder Network along with a Feature Network, which has handcrafted features for sentiment analysis of English Hindi code mix text.

\section{Embeddings}\label{section:embeddings}

In past few years, there has been a significant amount of work in bilingual, crosslingual and multilingual representation of text as they help us in building models for low resource languages. \cite{survery} presents a detailed analysis of cross-lingual word embedding models and discusses how lots of them optimize for the same objective function, and that different models are often equivalent. We choose the followings embeddings to work on our code-mixed task.

\subsection{Mapping based model}

In this approach, we first independently learn monolingual word representations, $L_1$ and $L_2$, from large monolingual corpus and then learn a transformation matrix $W$ to map representation from one language to the representation of the other language. The projection matrix transforms vector space of embedding of one language, $L_1$ to that of the other language, $L_2$. The projection matrix can be learnt in both supervised and unsupervised settings. \cite{conneau2017word} shows that the mapping learnt in an unsupervised settings can perform better to those learnt in supervised settings.

We have used MUSE\footnote{\url{https://github.com/facebookresearch/MUSE}} to obtain this mapping. This tool returns a word-aligned embedding space for each language, $L_1^{'}$ and $L_2^{'}$, such that similar words in either language have similar representation. We merge two different embedding space by taking an average of word vectors of common words and simply appending the others. 
As a result, we get a single embedding space, $L$, and vocabulary for both the languages. 

The monolingual embeddings for English and Spanish used in our experiments were obtained from \cite{grave2018learning}. MUSE provides both supervised and unsupervised variant. 

\subsubsection{MUSE-USUP}
It uses adversarial learning to learn the transformation matrix followed by refinement using Procrustes algorithm. Hence, this approach is independent of any kind supervised signals between two languages.

\subsubsection{MUSE-SUP}
It uses bilingual dictionary to learn a mapping from the source to the target space using Procrustes alignment. Its requirement of a bilingual dictionary makes it supervised approach.

\subsection{Pseudo-multi-lingual model}

In this approach, we obtain word/subword representation by training word embedding model, like fastText or Glove, on single corpus containing text from both the languages, usually formed by concatenation of monolingual corpus. This leads to formation of a shared vocabulary and all words vectors are in a common vector space. Words/Subwords from the two languages may or maybe not be aligned. The word embedding models chosen by us are subword based. Subwords allow us to guess the meaning of out-of-vocabulary words, which helps us to handle the large vocabulary and large amount unseen constructions caused by combining lexicon and syntax of two languages. Since most of the code-mixed text belongs to the category of texting language, there is a huge variations of spelling and misspelling. Subwords allow us to get proper word vectors for even misspelled words.

\subsubsection{MultiBPEmd}
\cite{heinzerling2018bpemb} released MultiBPEmd, a shared subword vocabulary and multilingual embeddings in 275 languages. It was formed by concatenating articles from all 256 languages on Wikipedia, learning a Byte Pair Encoding\citep{DBLP:journals/corr/SennrichHB15} subword segmentation model using SentencePiece\citep{kudo-richardson-2018-sentencepiece}, and using GloVe\citep{pennington-etal-2014-glove} to train subword embeddings. We use MultiBPEmd version with vocabulary size of 1,000,000 and dimensions of 300. 

\subsubsection{Custom-fastText}
fastText\citep{bojanowski2016enriching} is popular method for learning word embeddings. We choose it because it can create word vector for words absent from the trained embedding space, by summing up the character n-grams vectors. Therefore, it offer us advantages similar to that of subwords. 
We trained fastText on custom corpus generated by concatenating 70 millions English, 70 millions Spanish and 20 millions code-mixed tweets mined by us from twitter. Our embedding has vocabulary size of 1,343,436  and dimensions of 100.

\subsection{Sentence level multilingual autoencoder model}

Due to large amount of research done in Machine Translation, a huge amount of sentence-aligned parallel data is available for popular language pairs. Using this data and machine translation models, we can learn cross-lingual sentence-level representation. 

\subsubsection{LASER}

LASER\citep{laser} stands for Language-Agnostic SEntence Representations. It offers a shared multilingual sentence-level representation for 93 different languages. LASER consists of common language-agnostic Bidirectional LSTM encoder which provides the sentence embedding. The sentences embedding is then decoded by language-specific decoder. It's trained on 223 millions parallel sentences obtained by combining the OpenSubtitles2018, United Nations, Europarl, Tatoeba. Global Voices, and Tanzil corpus. LASER has a BPE vocabulary size of 50,000 and builds sentence representation of dimensions of 1024.

\section{Dataset}\label{section:dataset}

We have used a combination of monolingual and code-mixed data for our training and testing of our sentiment analysis model. All of our experiments use a combination of five different datasets listed below. 

\begin{itemize}
    \item \textbf{SemEval 2017}: We use 20,632 English tweets from dataset provided by SemEval-2017 Task 4:Sentiment Analysis in Twitter \citep{rosenthal-etal-2017-semeval}.
    \item \textbf{Sentistrength}: 4,241 English labelled tweets have been provided by Sentistrength\footnote{\url{http://sentistrength.wlv.ac.uk/}}.
    \item \textbf{TASS-2017}: We use 7,217 labelled Spanish tweets from \cite{Ceron17}.
    \item \textbf{TASS-2014}: We use 3,202 labelled Spanish tweets from \cite{tassa}.
    \item \textbf{Code-mixed Dataset}: \cite{en_es_datset} provides 2,449 train and 613 test instances of code-mixed tweets.
\end{itemize}

All of the instances have three labels: neutral, positive, and negative. Tweets have been prepossessed and tokenized using a version of TweetMotif\citep{TweetMotif} before going through embedding specific prepossessing.


\section{Training Curriculum}\label{section:training curriculum}


\cite{choudhury-etal-2017-curriculum} shows that mixing monolingual and code-mixed data randomly doesn't generate the best result for code-mixed models. Rather, first training the models with monolingual training instances and then training the resulting network on code-switched data is a better curriculum. Hence, we first train our models using instances from English and Spanish datasets combined and then, if required, we train the resulting model using instances from code-mixed dataset. The latter step is only used to get supervised results. This training curriculum is efficient for knowledge transfer from monolingual text to code-mixed text.


\section{Models}\label{section:models}

In this work, we perform sentiment analysis of code-mixed text. Our choice of this tasks is motivated by the availability of annotated code-mixed and monolingual dataset. Our task is to classify a code-mixed sentences as positive, negative, or neutral.  As our goal is to demonstrate effectiveness of embeddings, we use simple classifier for our task. Models using sentence embedding have a single layer dense neural network. Models using word/subword embedding have a single layer Bidirectional LSTM with 50 units. The LSTM layer has a dropout and recurrent dropout of 0.3. All of our model have an output layer of size of three and use Softmax. We use ADAM \citep{adam} optimiser with learning rate of 0.001 and momentum of 0.9. We use a batch size of 32. We use early stopping with the patience of 10. We use the model with best validation score at the current step to initialise the next step.


\begin{table}[!b]
\centering
\begin{tabular}{llll} 
\toprule
\textbf{ Model}  & \textbf{Embeddings }          & \textbf{Learning Type } & \textbf{F1-score }  \\ 
\toprule
\multicolumn{4}{l}{\textit{{\bf Previous Work}} }                                                      \\ 
\midrule
LSTM             & -                            & Supervised              & 54.40               \\
LSTM             & Bilingual Skip-gram\footnote{Reported in \cite{pratapa-etal-2018-word}}\        & Supervised              & 61.50               \\
GirNet\footnote{Reported in \cite{gupta2019girnet}}         & -                            & Supervised              & 63.40               \\
LSTM             & Synthetic CM based\footnote{Reported in \cite{pratapa-etal-2018-word}}           & Supervised              & \textbf{64.60}      \\ 
\midrule
\multicolumn{4}{l}{\textit{{\bf Our Work}} }                                                           \\ 
\midrule
LSTM             & MUSE-USUP                     & Supervised              & 55.40               \\
LSTM             & MUSE-SUP                      & Supervised             & 56.61              \\
LSTM             & MultiBPEmb                    & Supervised              & 64.00               \\
NN               & LASER                         & Supervised              & 64.44               \\
LSTM             & FastText                      & Supervised              & \textbf{67.71}      \\ 
\midrule
LSTM             & MUSE-SUP                      & Partially supervised    & 53.53              \\
NN               & LASER                         & Partially supervised    & \textbf{61.72}      \\ 
\midrule
LSTM             & MUSE-USUP                     & Unsupervised            & 51.99               \\
LSTM             & MultiBPEmb                    & Unsupervised            & 54.59               \\
LSTM             & FastText                      & Unsupervised            & \textbf{58.40}              \\
\toprule
\end{tabular}
\caption{F1-score of the competing models on code-mixed sentiment analysis task.}
\label{table:sa_result}
\vspace{-5mm}
\end{table}

\section{Results}

Table \ref{table:sa_result} reports the result of our experiments using embeddings in Section \ref{section:embeddings}, and models in Section \ref{section:models}, trained and tested using datasets in Section \ref{section:dataset} as per training curriculum in Section \ref{section:training curriculum}. 
We beat the previous state-of-the-art on the task of English-Spanish code-mixed sentiment analysis by an absolute 3.11\% using our custom-fastText embeddings. Custom-fastText embeddings gives us the best totally-unsupervised performance with 0.58 F1-score. This shows training on actual code-mixed corpus, and using subwords helps in developing better embeddings because just concatenated monolingual or parallel corpus fail to capture the fine syntactic and semantic features of code-mixed text. LASER produces a remarkable zero-shot F1-score of 0.62, and  performs at par with the previous state-of-the-art. This shows that training on translation tasks, with common encoder, and using parallel corpus indeed helps.

\section{Conclusions}

Today, code-mixing is a rapidly evolving way of expression in multilingual populations on social media and in-person conversation. Even the best monolingual sentiment analysis techniques perform poorly on code-mixed text because due to involvement of a new language and unseen structures in a single sentence. Here, we have shown that usage of off-the-shelf crosslingual and multilingual embeddings is an effective way to improve the performance of sentiment analysis techniques on code-mixed text. Since, crosslingual, in particular multilingual, research is progressing rapidly today, we can hope to see similar performance increase in code-mixed analysis domain. 
We believe that the future of code-mixed data analysis will lie in zero-shot approaches, where we efficiently transfer knowledge from monolingual datasets to code-mixed datasets using embeddings and models that can handle code-switching without much loss in performance.

\subsubsection*{Acknowledgments}

We're grateful to Divam Gupta for all his guidance throughout this project.

\bibliography{iclr2020_conference}
\bibliographystyle{iclr2020_conference}


\end{document}